\documentclass[letter, 10pt, conference, compsocconf]{IEEEtran}

\IEEEoverridecommandlockouts                              % This command is only needed if 

\usepackage{xcolor}
\usepackage{soul}
\usepackage{cite}
\usepackage{amsmath,amssymb,amsfonts}
\usepackage{algorithmic}
\usepackage{graphicx}
\usepackage{textcomp}
\usepackage{xcolor}
\usepackage{txfonts}
\usepackage{float}
\def\BibTeX{{\rm B\kern-.05em{\sc i\kern-.025em b}\kern-.08em
    T\kern-.1667em\lower.7ex\hbox{E}\kern-.125emX}}
\usepackage{subcaption}
\usepackage[colorlinks]{hyperref}
\usepackage[normalem]{ulem}

\begin{document}

\title{Mobile Robot Control and Autonomy Through Collaborative Simulation Twin}

\author{Nazish Tahir \and Ramviyas Parasuraman 
\thanks{The authors are with the Heterogeneous Robotics Research Lab, School of Computing, University of Georgia, Athens, GA 30602, USA. 

Emails: \tt{nazish.tahir,ramviyas\}@uga.edu}}}

\maketitle

\begin{abstract}
When a mobile robot lacks high onboard computing or networking capabilities, it can rely on remote computing architecture for its control and autonomy. This paper introduces a novel collaborative Simulation Twin (ST) strategy for control and autonomy on resource-constrained robots. The practical implementation of such a strategy entails a mobile robot system divided into a cyber (simulated) and physical (real) space separated over a communication channel where the physical robot resides on the site of operation guided by a simulated autonomous agent from a remote location maintained over a network. Building on top of the digital twin concept, our collaborative twin is capable of autonomous navigation through an advanced SLAM-based path planning algorithm, while the physical robot is capable of tracking the Simulated twin's velocity and communicating feedback generated through interaction with its environment. We proposed a prioritized path planning application to the test in a collaborative teleoperation system of a physical robot guided by ST's autonomous navigation. We examine the performance of a physical robot led by autonomous navigation from the Collaborative Twin and assisted by a predicted force received from the physical robot. The experimental findings indicate the practicality of the proposed simulation-physical twinning approach and provide computational and network performance improvements compared to typical remote computing (or offloading), and digital twin approaches.

\end{abstract}


\section{Introduction}

Remote operation allows a safe way to control a robotic system, avoiding difficult or harmful environments that may be distant or out of reach for humans. %Teleoperation builds a bridge between the control, perception, and cognitive capabilities of humans to the precision, robustness, and mobility of robots. 
Mobile robots require robust control and autonomy algorithms as well as cooperation algorithms to be useful in applications like exploration \cite{muscolo_marcheschi_fontana_bergamasco_2021} and search and rescue missions \cite{Nourbakhsh2005,yang2020needs}, where human access is difficult, but supervised human or artificial intelligence control is necessary \cite{nguyen2001virtual}.
Further, remote control and autonomy add the possibility of enabling an operator/AI algorithm to remote interface with the on-site environment robot by effectively using multi-modal sensor data inside the loop such as visual \cite{Kelly2011}, tactile-visual \cite{Mut2002}, visual-vestibular \cite{RobuffoGiordano2010}, haptic cues \cite {Lee2011a,Son2013}, artificial force  \cite{Lam2007}, vibrotactile \cite{Scheggi2014}, visual-network \cite{parasuraman2017new}, etc. This remote operation can be extended to collaborative robots to control a remote robot safely.

\begin{figure}[t]
    \centering
    %\vspace{-1cm}
    \begin{center}
    \includegraphics[width=.99\linewidth]{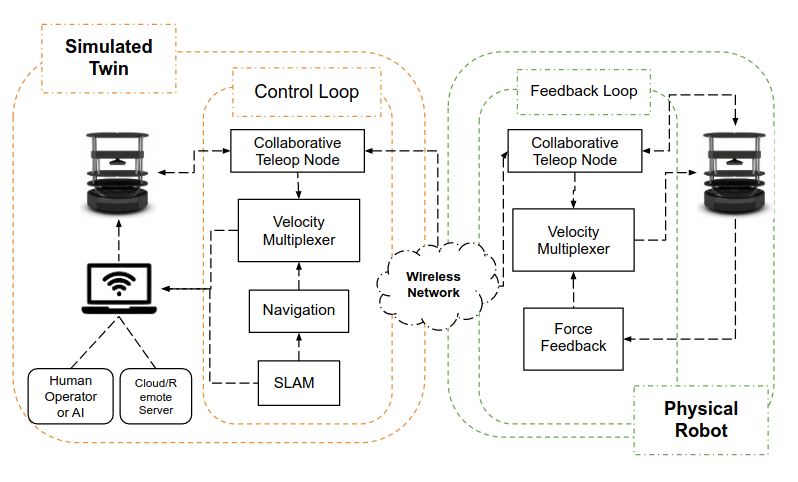}
    \caption{An overview of the collaborative Simulation Twin (ST) architecture with a Simulated-Physical robot collaboration for mobile robot teleoperation and autonomous navigation.}
    \label{fig:main-fig}
    \vspace{-6mm}
\end{center}
\end{figure} 

%In this context, a robot remote teleoperation comes into play where a human operator commands the remote robot through an interface and receives informative feedback cues to gain situational awareness of the remote site. 

Digital Twin (DT) \cite{LU2020101837} and simulation-based approaches play an important role in robotics for unit testing and validation of control schemes or algorithms, mainly during development.
A DT is mostly applied as a training and control technique for pre-testing purposes when the simulators or virtual interfaces are provided with a model of the robot that may imitate the actual behavior of the real robots as precisely as possible.  Simulation models allow for the performance assessment in a completely safe environment while considerably speeding up the optimization process \cite{silva2016open,PHANDEN2021174}. 

%Unlike DTs, which are exclusive to the concept of pre-testing and training for educating robots, one possible approach to using simulated models is to intuitively use them for the direct control of tasks like teleoperation and coordination of robots to reduce redundant learning costs which leads to unpredictability associated with training or learning processes that arise due to intrinsic mismatch between simulated testing environment when transferred to the real world \cite{PHANDEN2021174}. 
Building on the DT concept, simulation models can be used to direct control of tasks like navigation or autonomy decisions and coordinating robots. 
Such a paradigm will help reduce redundant learning costs, leading to unpredictability associated with training or learning processes that arise due to intrinsic mismatch between simulated testing environments when transferred to the real world \cite{PHANDEN2021174}. 
This paradigm also involves robot control in a physically secure and technologically adaptable space while bridging the mismatch between the simulation and the real environment, thus helping analyze the simulation to reality gap \cite{salvato2021crossing}. Its usefulness in collaborative robot systems, particularly if a human expert or a simulated AI agent is present in the loop to control the robot remotely rather than train it, appears promising. 
This enables remote control of the robot by working collaboratively with a human or a simulated AI agent to do routine tasks, eliminating extra workload on the robots and increasing efficiency.  

%Working collaboratively, remote teleoperation of the robots relies on the high-level cognitive abilities of a human operator or an autonomous AI agent to completely navigate a complex environment. The major task in creating safe and productive cooperation between the operator and robot is to build an intuitive connection by linking the simulator and the robots for direct control of robots without resorting to specialized visual technologies. 

Therefore, we propose a novel remote control and collaboration framework through the concept of a collaborative Simulated Twin (ST), which introduces a collaboration layer with a DT and leverages the computing capabilities of the simulation twin for the physical robot's autonomy and navigation, enhanced with reduced networking requirements compared to typical computational offloading. 
%Using a high-fidelity virtual model with physical world geometry, features, behavior, and data, a Simulated Twin aids in integrating a real robot with a simulated virtual robot counterpart to achieve precise sensory control and improve system performances for motion, navigation, and orientation.  
The overall framework of the proposed ST-enabled robot collaborative teleoperation system is depicted in Fig.~\ref{fig:main-fig}.

%A competent simulation system models an ST dwelling in a remote location, allowing highly synchronized monitoring and teleoperation of an actual field robot. The physical robot's movement can be controlled remotely using the mapped ST of the physical robot, and the physical robot's status may be monitored and viewed using the robot ST, thus resulting in higher situational awareness by the operator of the robot states. Furthermore, by incorporating the proposed prioritized-based switching control approach via a communication mechanism, robots may interact even when deployed in separate locations, which is consistent with collaborative robot teleoperation. 

Robot collaboration is achieved by the ST taking over more powerful or computationally heavy tasks from the physical robot, resulting in fewer processing deadlocks at the operation site. In response, the physical robot establishes a feedback loop with the ST to maintain flawless synchronization between the actual and simulated models, thus mimicking the real system as precisely as possible.

The key novelties proposed in our approach are as follows. 
\begin{itemize}
    \item A simulated-physical robot collaboration framework is introduced, with the possibility to integrate human intervention. The framework uses computational offloading and collaborative autonomy to optimize the network and computing performances.
    \item To guarantee better goal point accuracy and obstacle avoidance, we incorporate a feedback mechanism that, instead of being applied to a joystick as is done traditionally in a control system, is applied to an autonomous robot located on the remote ST.
    \item We are interested in stabilized communication between ST and the physical robot. The data storage and processing should be able to handle massive amounts of data, as well as intermittent connectivity or data loss. Hence, we explore the computational and network performance of the proposed system across multiple scenarios of increasing complexities.
\end{itemize}

This work is an extension of a physical twinning architecture introduced in our recent works \cite{tahir2022analog,tahir2020robot}, where two robots are tightly coupled in terms of their control loops to exploit the advantages of collaborative computing and networking. As a complementary work to \cite{tahir2022analog}, this paper provides a new perspective of a simulation twin controlling a real robot. Our objective is to capture the interplay between the simulation-reality control gap, which can inform future collaboration architectures employing both simulation systems and real-world robots inside the control and autonomy loop.

\begin{figure*}[t]
    \centering
    %\vspace{-1cm}
    \begin{subfigure}{.24\textwidth}
        \includegraphics[width=\textwidth]{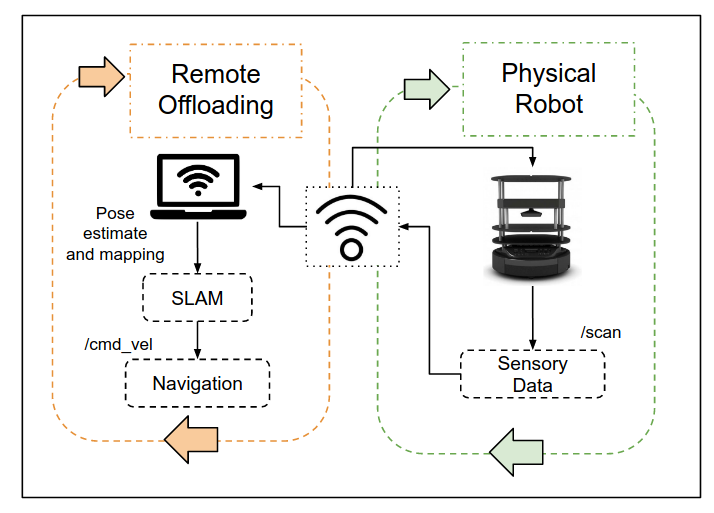}
        \caption{Remote Offloading (RO)}
        \label{fig:RO}  
        %\vspace{-5mm}
    \end{subfigure}
    \begin{subfigure}{.24\textwidth}
    %\vspace{-1cm}
        \includegraphics[width=\textwidth]{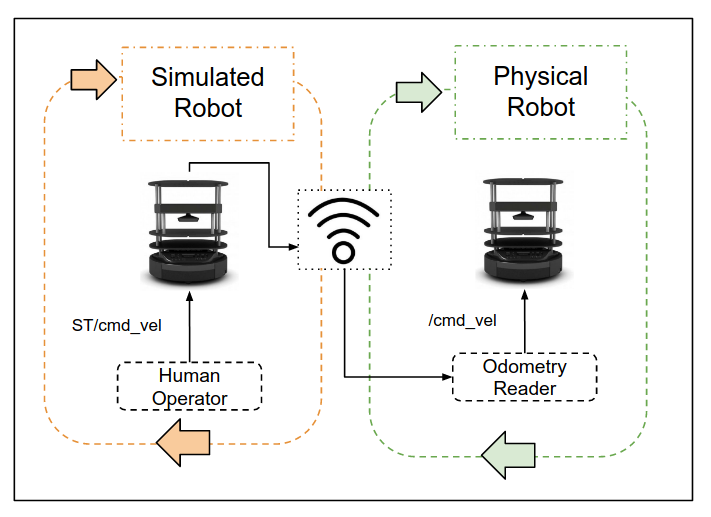}
        \caption{Manual Teleoperation (MT)}
        \label{fig:MT}  
    %\vspace{-5mm}
    \end{subfigure} 
    \begin{subfigure}{.24\textwidth}
    %\centering
    %\vspace{-1cm}
        \includegraphics[width=\textwidth]{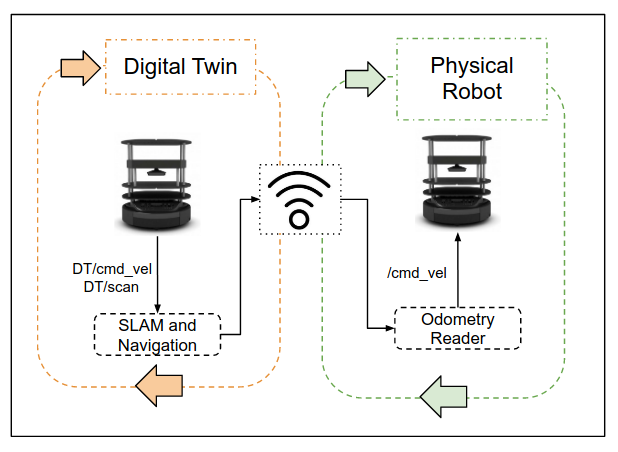}
        \caption{Digital Twin (DT)}
        \label{fig:DT}  
    %\vspace{-5mm}
    \end{subfigure}
    \begin{subfigure}{.24\textwidth}
    %\centering
    %\vspace{-1cm}
        \includegraphics[width=\textwidth]{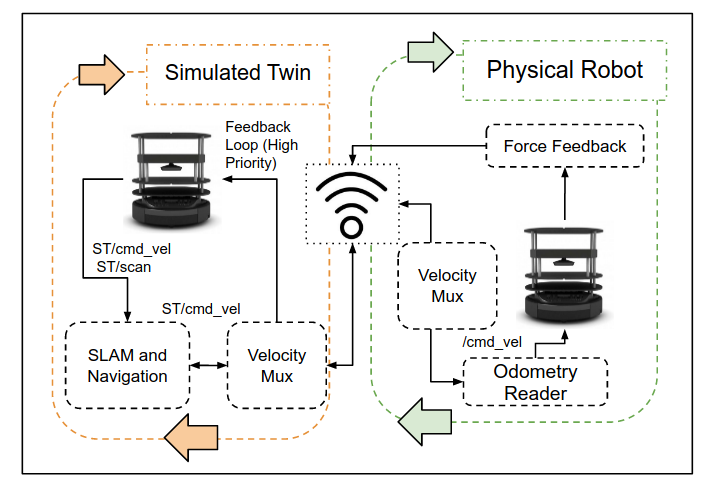}
        \caption{Proposed Simulated Twin (ST)}
        \label{fig:ST}  
    %\vspace{-5mm}
    \end{subfigure}
    %\vspace{-2mm}
    \caption{Thematic representation of the experiment cases analyzed in this experimental study. Here, 'odom', 'scan', and 'cmd\_vel' are the ROS topics for odometry, LIDAR scan, and command velocity (control) information.}
    \label{fig:cases}
    %\vspace{-5mm}
\end{figure*} 

\section{Related Work}
The concept of a digital twin has been applied in literature for a range of applications such as Smart Manufacturing \cite{david2019unified}, monitoring and management \cite{tao2018digital}, and virtual robot behavior adaption \cite{kousi2019digital}.  The DT has been widely used in manufacturing systems' design and performance assessment to allow dynamic updating of the twin with data about the actual production process from multiple sensors\cite{kousi2021digital}. It focuses on a model-based simulation that combines manufacturing processes with the full product life cycle and represents the next wave in manufacturing modeling, simulation, and optimization \cite{app12104811}. Recently, its usage has been established to allow remote control in production systems, e.g., the building of a DT of the robot arm in the Unity engine to acquire manufacturing skills digitally, where the real robot arm may then reproduce the learned abilities in physical space \cite{matulis2021robot}. In a similar fashion, authors of \cite{DBLP:journals/corr/abs-2103-10804} created an immersive human-in-the-loop robotic assembly system by combining DTs with virtual reality (VR) interfaces.

%Robot control and teleoperation entail the user controlling the robot from a distance via an appropriate interface (joystick, keyboard, etc.). The robot teleoperation control paradigm, which benefits from a strong coupling of human input to robot actions, is a well-studied field of research in robot control. Many technologies have emerged in recent years to enable remote robot teleoperation through virtualization. 
Simulation techniques have been applied to robotics via virtual testbeds \cite{cichon2017robotic}, in which intricate technological systems and their interactions with potential working environments are created, programmed, controlled, and optimized in 3D simulation before commissioning the real system. In \cite{cichon2016simulation}, the use of simulation in conjunction with mental models is used as a conceptual extension of a virtual testbed toward simulation-based control and simulation-based assistance. Stepping into sophisticated simulated interfaces, \cite{ong2020augmented} proposed an augmented reality-based robot programming system that converts the robot's work environment into AR scenes, enabling rapid and intuitive robot route planning and task programming. The authors of \cite{sadeghi2016cad2rl} train a vision-based navigation strategy exclusively in simulation before attempting to apply it to a quadrotor without attempting further training runs with deep reinforcement learning. Another work \cite{segura2020visual} used visual computing techniques to use partially simulated input on the physical environment in real time by having an operator virtually interact with the collaborative robot.

% DT in HRC 
 %\textcolor{red}{The application of DT in the design and control of Human-Robot Collaborative (HRC) activities has also been researched \cite{bilberg2019digital}. Another paper \cite{DRODER2018187} proposes a machine learning enhanced DT approach for the safe movement of the robot in a human-robot collaboration setting.}

A DT has been employed to perform remote robot teleoperation where the human operator controls the DT, while DT controls the remote robot \cite{9013428}. 
%The human-machine-human control loop is divided into Virtual Reality (VR), remote control, and robot control loops. 
The suggested architecture enables low-latency visual feedback and very quick system reaction times for unanticipated changes with arbitrary distances between the operator and robot.
In a more recent work \cite{LI2022102321}, a multi-robot collaborative production system with human-in-the-loop control using augmented reality (AR) and DT technologies. The DTs of industrial robots are first mapped to real robots and then seen in AR glasses in their proposed technique, and a multi-robot communication system is built to synchronize the states of the robots in the twin. A reinforcement learning approach is also used in robot motion planning.

However, we depart from the vision of the DT as several discrepancies arise between the DT's forecast and the physical asset's process data, which leads to a simulation-reality gap. DTs may require different modeling methods and additional tools that may need to be developed and integrated for them to be implemented successfully, which requires added cost and effort. Authors of \cite{shao2018digital} highlight several issues and challenges associated with using DTs in the Cyber-Physical Systems (CPS) domain. With the use of Simulated Twin (ST), we remove any need for learned behavior as we focus on a constant mutual interaction between the cyber and physical world through the control loops. Instead of training and testing first on simulation and transferring the results to reality, in the case of ST, simulation and physical execution coexist. Therefore, a transfer to reality and a transfer back are required. Specifically, compared to the DT literature, we depart by enabling the tight integration of the simulation-based system (which can operate in a different environment) with the real system's control and autonomy for optimizing network and computational offloading. The ST architecture also allows the simulated-physical spaces to significantly deviate from each other, unlike DT architecture.

\section{Proposed Approach}
To remotely teleoperate a mobile robot, we propose the idea of a simulated twin (ST) which ensures effective communication with the physical robot. %We specifically test our strategy in a remote teleoperation domain. 
Currently, smart manufacturing systems performing remote teleoperation do not directly involve human-in-the-loop control approaches and automated control strategies are encouraged \cite{lim2020state}. To meet a similar idea, we propose a completely autonomous ST that remotely controls a physical robot over a communication channel. Firstly, a mechanism of Collaborative communication is established between the ST and the Physical Robot. Next, prioritized-based switching control is applied, which allows constant switching between simulated automated control to allow adjustments for the real-time incoming stream of data from the physical robot and its environment. 
Fig.~\ref{fig:cases} provides an overview of the proposed approach compared to existing approaches, which are later detailed in Sec.~\ref{sec:cases}.

\subsection{Remote Robot Control} 
We introduce the concept of a simulated robot (ST) paired over a communication medium with a physical robot divided into Cyberspace and Physical space, respectively. The ST acts like a virtual surrogate to the physical robot in a Simulated-Physical relationship and is capable of performing autonomous navigation in its environment, while the physical robot present at a remote location can replicate the commands given to the ST. To operate in an unknown environment, the robot needs to build a map of the unexplored environment, a computationally expensive task offloaded to the surrogate ST. 

Simulated Twin creates a robust and accurate ambient map using SLAM (Simultaneous Localization and Mapping \cite{thrun2008simultaneous}) while also localizing itself inside that map. The robot must maintain two types of representations at the same time: an environment observation model and a localization model or location estimation. That is, they should be executed iteratively, with one output being the input of the other. While navigating uncharted territory, the robot creates a 2D occupancy grid map using SLAM, in which the environment is represented in a discrete grid of rectangular cells of the same size and shape. Each cell is allocated a value that represents the likelihood of occupancy. 

We use the occupancy grid map thus obtained to perform path planning using Dynamic Window Approach (DWA). DWA is a well-known obstacle avoidance and navigation method that works well in experimental settings. DWA considers a mobile robot's dynamic and kinematic restrictions. It enables the robot to come to a halt before colliding with an obstruction and performs an optimization that determines the quickest, unobstructed path to the target. It also brings the robot to a halt in the desired target location and executes a recovery rotation.

ST performs navigation by moving towards a certain target point on the observation map, either instructed by an operator (human-in-the-loop) or through a waypoints navigation script (autonomous task) run at the control site. Since the Physical twin is coupled with the ST and the environment in Cyberspace is closely modeled according to the physical world, the physical twin replicates the commands issued at the ST and performs navigation at the remote location.

\begin{figure*}[ht]
\centering
\begin{subfigure}{.28\textwidth}
\centering
\includegraphics[width=\linewidth]{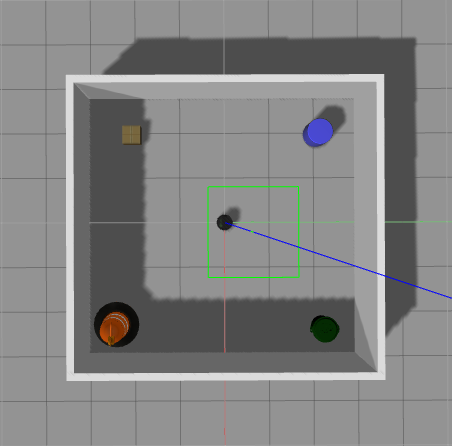}
\caption{Simulated Robot Twin}
\label{fig:simulated_env}
\end{subfigure}
\begin{subfigure}{.37\textwidth}
\centering
\includegraphics[width=\linewidth]{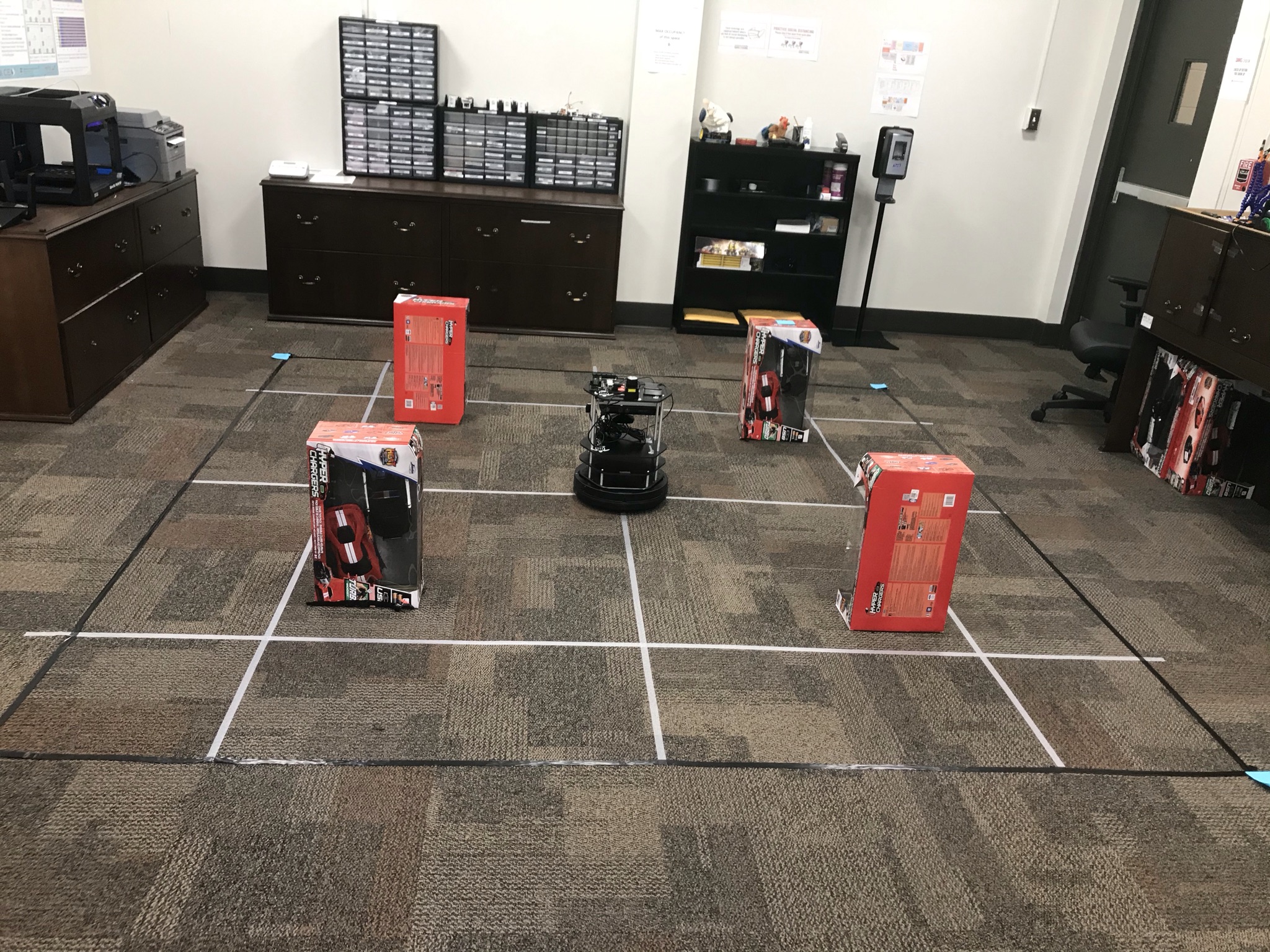}
\caption{Physical Robot and Environment}
\end{subfigure}
\label{fig:physical_env}
\begin{subfigure}{.32\textwidth}
\centering
\includegraphics[width=\linewidth]{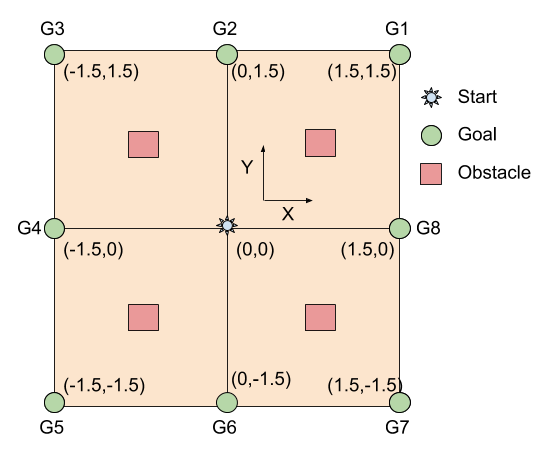}
\caption{Goal locations for navigation}
\label{fig:goal_points}
\end{subfigure}
%\vspace{-2mm}
\caption{Experimental setup of Simulated-to-Physical robot collaboration using Simulation Twin.}
\label{fig:Sim2real_setup}
\end{figure*}

\subsection {Multi-robot Collaboration}
SLAM techniques are computational and data-intensive algorithms that are challenging to implement on smaller, less costly robots that are typically used in the field to keep costs down.
By efficiently offloading these algorithms to an autonomous agent, such robots gain a substantial advantage in executing tasks remotely without incurring operating costs. By replicating the Simulated Twin's commands and offloading computationally intensive tasks like navigation, the physical robot saves a significant amount of processing and memory resources. Also, communication and state synchronization between the Simulated Twin and the physical robot is the backbone of our proposed strategy. To keep the simulated robot's states precisely synced with the real robot, a robust and reliable network setup is required, which is why we are interested in the suggested framework's reliability under realistic network settings. In this research, we utilize the widely-used robotic middleware framework -- Robot Operating Systems (ROS) \cite{quigley2009ros} -- to publish and subscribe to topics over Simulated and Physical robots to maintain network connectivity over a WiFi channel. The physical robot's movement is controlled remotely using the mapped ST, so that its status may be monitored and viewed using ST in the simulated space. Because the physical robot mimics ST's movements, the trajectories may be visualized on the simulator to avoid any safety hazards. Further, the use of simulation provides a user-friendly way for the operators to view the physical robot's spatial movements.

\subsection{Prioritized Switching-Based Control} 

A simultaneous bidirectional link is maintained on both sides of our proposed prioritized Simulated-to-Physical robot control strategy through a low-priority queue velocity multiplexer (MUX). MUX's principal job is to arbitrate incoming velocity commands over several channels, enabling only one topic to publish at a time. A collaborative coupling is established by allowing the Simulated robot to take Human/AI's velocity commands on a low priority channel, while the physical robot subscribes to the ST's velocities on a low-priority channel as well. Both agents have simultaneously subscribed to the reactive feedback force on a high-priority channel.

Our proposed strategy involves a simulated robot (ST) navigating its environment while the physical robot subscribes to the ST's state to mimic its mobility at the field location. However, the physical robot has an additional layer of safe navigation for reactive obstacle avoidance given by a quick, low-intensive predictive force feedback calculation communicated to the ST over a high-priority channel. The path must be readjusted after applying force feedback to the ST for safer navigation around a complex environment at the physical robot's site. The ST offers predictability for the physical robot's ultimate posture and motion trajectories, besides offering a way for the physical robot to avoid any safety hazards through the employed feedback loop.

\section{Experimental Analysis}
Here, we present the proposed framework's experimental results and compare them to existing approaches. Several target points were tested for transfer to a collaborative physical robot to test the remote teleoperation scheme proposed through a simulated robot to ensure the efficacy and practicality of the proposed framework.
The network and computational parameters were calculated while experiments were performed for three baseline experimental cases against the proposed strategy (Fig. \ref{fig:cases}). The experiment setup and the target points examined are illustrated in Fig. \ref{fig:Sim2real_setup}. %namely (1.5,1.5), (1.5,-1.5), and (1.5, 0).

\paragraph*{Simulated Robot} 
On a machine running Ubuntu 18.04 and supporting ROS melodic, a simulated robot is spawned in the Gazebo simulator. The world file in Gazebo is a 6x6m walled area with obstacles added at the corners to assist map out the empty space. The model of Turtlebot 2e is employed as a Simulated Twin for the physical Turtlebot 2e robot. The computer spawning the simulated model is around 20 meters distant from the real robot, which is located in another room. 

\paragraph*{Physical robot} 
The physical system is a Turtlebot 2e robot equipped with a 2D laser rangefinder and an Nvidia Jetson Nano as the onboard CPU. It's in a 3x3m space with obstacles at each corner. The real robot connects to the same WiFi network as its ST in the lab.

\begin{figure*}[ht]
\centering
\begin{subfigure}{.32\textwidth}
\centering
\includegraphics[width=\linewidth]{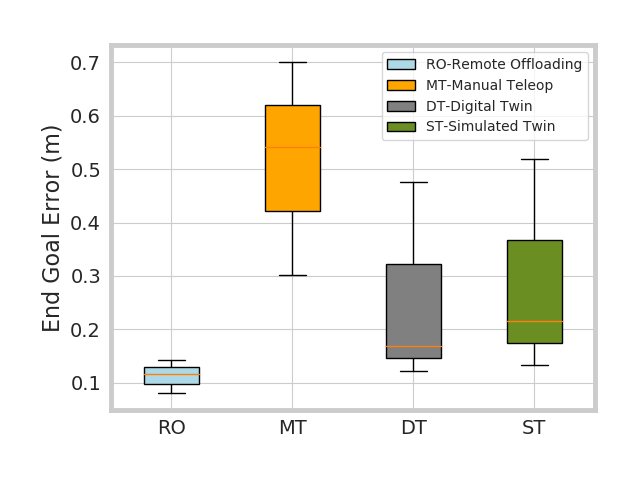}
\vspace{-8mm}
\caption{Goal Accuracy}
\end{subfigure}
\begin{subfigure}{.32\textwidth}
\centering
\includegraphics[width=\linewidth]{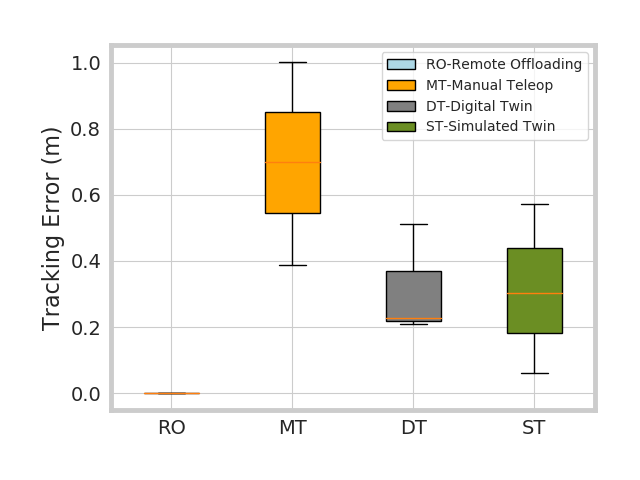}
\vspace{-8mm}
\caption{Tracking Accuracy}
\end{subfigure}
\begin{subfigure}{.32\textwidth}
\centering
\includegraphics[width=\linewidth]{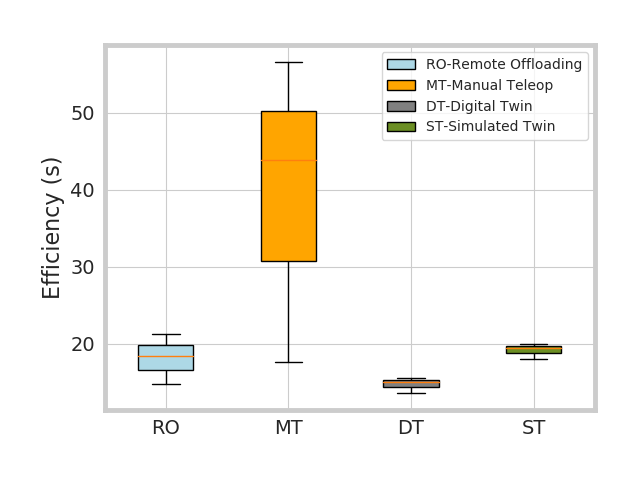}
\vspace{-8mm}
\caption{Efficiency}
\end{subfigure}
\caption{Task Performance: End Goal, Tracking Error and Efficiency in Collaborative Twin robot framework experiments.}
\label{fig:s2r-goal-plots}
\end{figure*}

\subsection{Experimental Cases}
\label{sec:cases}
%The following experimental cases are considered. 

\paragraph*{1. Remote Offloading (RO)} We test and evaluate our proposed strategy against this conventional baseline case where a hardware robot situated on a physical field utilizes the powerful computational capabilities of a remote server, breaking its dependency on onboard processing.  A typical offloading scenario is constructed where remote intelligence is applied by separating a robot physically and logically. The computing responsibilities of the robot are offloaded to a distant server or a Cloud service, and the robot's operation is managed remotely by either an AI agent or a human operator. In this scenario, the goal is sent to the physical robot on the field via the network from the output of the SLAM mapper and path planner. In this experiment, no simulated twin is employed, but a remote computer/laptop is used for control. This instance enables the physical robot to delegate computationally heavy tasks like SLAM, path planning, and navigation to a remote server/agent.

\paragraph*{2. Manual Teleoperation (MT)} A conventional teleoperation system with a human-in-the-loop is employed, where a human operator teleoperates the Physical robot through the simulated robot. The human operator sits at the control station and teleoperates the simulated model to the desired goal point using a gamepad in the simulated world, while the Physical robot reads the odometry commands through ROS topics to achieve the same target location as the simulated robot.
    
\paragraph*{3. Digital Twin (DT)} We recreate the scenario of a DT in this instance to test against the proposed strategy. The DT of the robot is synchronized with the real robot as a virtual model in the simulator Gazebo's workspace. The movement of the physical robot may be tracked and viewed by the robot DT using the mapped DT of the physical robot. Here the simulated workspace, as well as the physical workspace, are devoid of obstructions and the case is solely evaluated for the direct control of the robot. The DT employs SLAM and navigation stack and reads the waypoints through the script to navigate the physical robot to the targeted poses. The DT offers predictability for the physical robot's ultimate posture and motion trajectories. Because of mimicking the DT's movements, the trajectories of the real robot may be viewed through DT to avoid any safety hazards. 

\paragraph*{4. Simulated Twin (ST)} The proposed technique is tested on a physical robot in a 3x3m square at a physical/field site, coupled with its simulated twin (ST) at a distant site using Gazebo Simulator in ROS. We place hurdles in the field to enhance complexity and put our method to the test in increasingly difficult scenarios while the simulated workspace is kept unaltered. The simulated and physical robots maintain a bidirectional connection by employing the use of the aforementioned velocity multiplexer. The physical robot offloads its challenging computational tasks onto ST which employs the SLAM and navigation stack to maintain map and path planning for arriving at the targeted goal points. In this case, the physical robot also generates a predictive feedback force in response to obstructions in its workspace, which is directed to the ST via the network. Upon receiving this feedback on its highest priority MUX channel, the ST reroutes and optimizes its trajectory based on the new information to arrive at the target goal point.

\begin{figure*}[ht]
\centering
\begin{subfigure}{.32\textwidth}
\centering
\includegraphics[width=\linewidth]{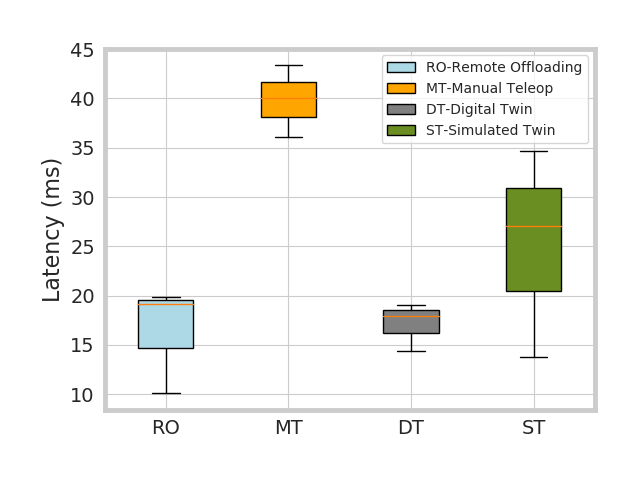}
\vspace{-2mm}
\caption{Network Latency}
\label{fig:s2r-latency}
\end{subfigure}
\begin{subfigure}{.32\textwidth}
\centering
\includegraphics[width=\linewidth]{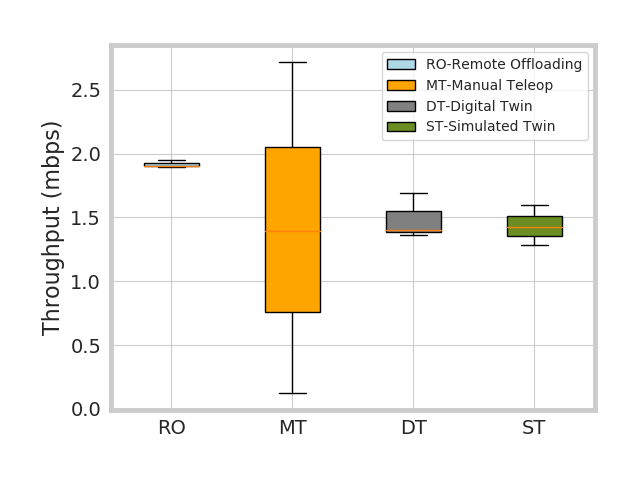}
\vspace{-2mm}
\caption{Throughput}
\label{fig:s2r-throughput}
\end{subfigure}
\begin{subfigure}{.32\textwidth}
\centerline{\includegraphics[width=\linewidth]{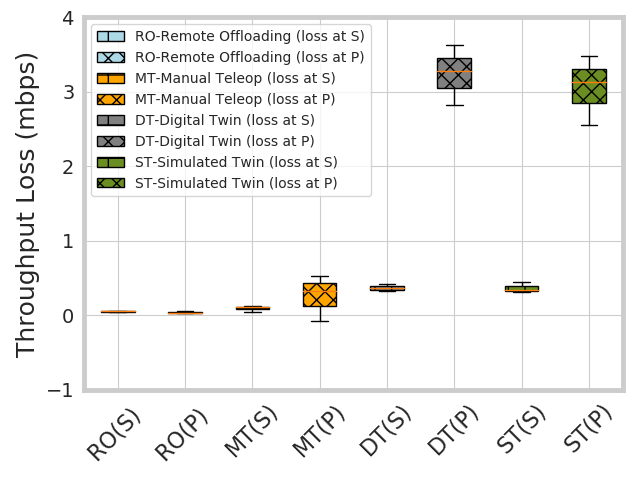}}
\vspace{-2mm}
\caption{Throughput Loss}
\label{fig:s2r-throughput_loss}
\end{subfigure}
\caption{Network Performance: Delay, throughput, throughput loss in Simulated Twin robot framework experiments. In 5(c), S denotes the Simulated site while P denotes the Physical site.}
\label{fig:s2r-network-plots}
\end{figure*}

\begin{figure*}[ht]
\centering
\includegraphics[width=0.45\linewidth]{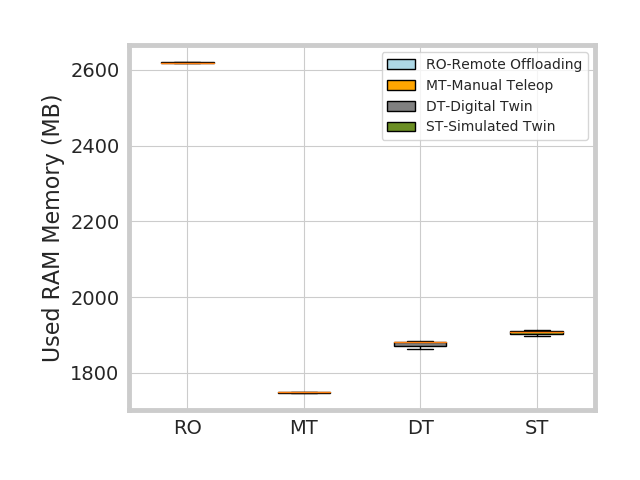}
%\caption{Computing Performance: Memory Utilization in Simulated Twin robot framework experiments.}
%\label{fig:s2r-computing-memory-plots}
\includegraphics[width=0.45\linewidth]{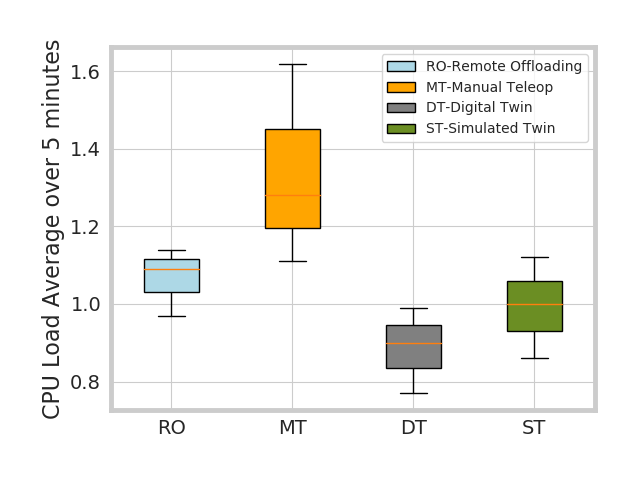}
\caption{Computing Performance: CPU and RAM Memory utilization in Simulated Twin robot framework experiments.}
\label{fig:s2r-computing-plots}
\end{figure*}

\subsection{Performance Metrics}
\label{sec:metrics}
%TODO - List the performance metrics as an enumerated list so it's easy to read and follow
We evaluate the system using performance metrics that quantify the navigation task as well as network and computing.

Task performance metrics include task \textit{goal accuracy} which entails the absolute difference between the final robot position and the desired goal position given, \textit{tracking accuracy} by taking an average of the absolute difference between the instantaneous positions of the simulated and the physical robots and task \textit{efficiency} through total time elapsed to reach a goal point by the physical robot.

Network performance metrics measure \textit{network latency} which is the average delay with which the simulated and the physical machines communicate, \textit{network throughput} which is the average data rate at which the ST-Physical robots communicate, and \textit{network throughput loss} which is the difference between the throughput obtained between the Simulation and the Physical sides. It is indicative of the packet loss in the network to estimate how much of the offloaded data was lost in the network.

Computing performance metrics entail the  percentage of \textit{CPU utilization} which is the average CPU load (measured every 5 minutes) at the physical robot during the experiments
and \textit{memory utilization} in MBs indicating the average RAM memory used by the physical robot during the experiments.

\subsection{Results and Discussion}
Below we detail the task performance, networking, and computing improvements achieved through the proposed ST.

\paragraph{Performance-related measures}
Fig. \ref{fig:s2r-goal-plots} shows the performance of the proposed methodology in a Simulated-to-Physical robot setup. All the trials were successfully conducted through the collaborative coupling of mobile robots, and both agents were able to navigate toward the goals given.

We discover that the target error is substantially higher in our suggested technique than in RO since we are now mapping states from the simulated robot to the actual robot, as opposed to RO, where the robot simply offloads its computing responsibilities to the Cloud or remote server. Such a result is largely expected by the delay experienced by ST; yet, the difference in the goal inaccuracy between DT and ST is not considerable. The proposed strategy also outperforms Manual Teleoperation (MT) as it takes longer for the human operator to find an optimum path toward the target position through the gamepad. In terms of tracking inaccuracy, there is no ST involved in RO, therefore there are no appropriate grounds for comparison, while DT and ST indicate a significant difference in tracking due to the introduction of obstacles in the environment. As with the efficiency of task completion, adopting ST in obstructed scenarios resulted in a success rate of 64\% in completing tasks compared to traditional RO.

\paragraph{Network-related measures}
According to the data in Fig.~\ref{fig:s2r-throughput}, the Manual Teleoperation (MT) and Remote Offloading (MO) cases received the highest throughput, whereas DT and ST experienced the most loss at the physical robot. This increase in throughput at MT is attributable to the lowest throughput loss according to Fig.~\ref{fig:s2r-throughput_loss}.
Similarly, the observed network latency at ST was almost two times that of RO. Queuing on intermediary network devices might impact data latency which in the case of ST is most utilized per Fig.~\ref{fig:s2r-latency} There is a variation across all cases concerning delay experienced however, MT's case experienced the most delay and highest throughput.

\paragraph{Computing-related measures} 
The performance of the Simulated-to-Physical robot regarding memory usage and CPU usage is shown in Fig. ~\ref{fig:s2r-computing-plots}. Memory usage for cases ST is less than that of traditional offloading (RO) and the observed trend for CPU usage considering uptime averaged over 5 minutes shows higher values compared to traditional RO and MT which validates the proposed technique's optimal usage of computational resources.

\section{Conclusion} 
In this paper, a novel Simulated Twin (ST) based remote robot teleoperation and collaboration control are proposed that does not involve human-in-the-loop and compared against baseline teleoperation cases of remote offloading, simple teleoperation through simulated twin, and a digital twin. 
Through robot-robot collaboration, the on-field robot can be simple in design and computation capacity as it relies on a heavier, more computational-backed ST for its computational tasks and presents a novel approach to offloading. 
The ST predicts the physical robot's eventual states and motion trajectories, while also allowing the physical robot to avoid any safety issues via the proposed feedback loop. 
The performance, network, and computational challenges are reviewed and tested through experiments, which are then compared to baseline situations to assess feasibility. 
The results show that our suggested approach has a high potential for generalization over a wide range of application scenarios.
We anticipate that the output of this technique will aid in the creation of a control interface for mobile robots that can adapt between several control modes, ranging from manual teleoperation to full autonomy.

\bibliography{simulation_twin}
\bibliographystyle{IEEEtran}
\end{document}